\documentclass{article}

\usepackage{microtype}
\usepackage{graphicx}
\usepackage{subcaption}
\usepackage{booktabs} 

\usepackage{hyperref}

\usepackage[accepted]{icml2024}

\usepackage{amsmath}
\usepackage{amssymb}
\usepackage{mathtools}
\usepackage{amsthm}
\usepackage{hyperref}
\usepackage{listings}

\usepackage[capitalize,noabbrev]{cleveref}

\theoremstyle{plain}

\theoremstyle{definition}

\theoremstyle{remark}

\usepackage[textsize=tiny]{todonotes}

\icmltitlerunning{Noisy Algorithmic Chain of Thought}

\begin{document}

\twocolumn[
\icmltitle{Understanding the Effect of Noise in LLM Training Data \\ with Algorithmic Chains of Thought}





\icmlsetsymbol{equal}{*}

\begin{icmlauthorlist}
\icmlauthor{Alex Havrilla}{gatech}
\icmlauthor{Maia Iyer}{ibm}
\end{icmlauthorlist}

\icmlaffiliation{gatech}{Department of Mathematics, Georgia Institute of Technology, Atlanta, USA}
\icmlaffiliation{ibm}{IBM Research, Yorktown Heights, USA}

\icmlcorrespondingauthor{Alex Havrilla}{ahavrilla3@gatech.edu}

\icmlkeywords{Machine Learning, ICML}

\vskip 0.3in
]

\fancyfoot[C]{\thepage}



\printAffiliationsAndNotice{\icmlEqualContribution} 

\begin{abstract}

During both pretraining and fine-tuning, Large Language Models (\textbf{LLMs}) are trained on trillions of tokens of text of widely varying quality. Both phases of training typically involve heuristically filtering out ``low-quality'' or \textit{noisy} training samples, yet little is known quantitatively about how the type or intensity of noise affects downstream performance. In this work, we study how noise in chain of thought (\textbf{CoT}) impacts task performance in the highly-controlled setting of algorithmically solvable tasks. First, we develop the Traced Integer (\textbf{TInt}) framework to generate highly customizable noised execution traces for any arithmetic function on lists of integers. We then define two types of noise: \textit{static} noise, a local form of noise which is applied after the CoT trace is computed, and \textit{dynamic} noise, a global form of noise which propagates errors in the trace as it is computed. We then evaluate the test performance of pretrained models both prompted and fine-tuned on noised datasets with varying levels of dataset contamination and intensity. We find fine-tuned models are extremely robust to high levels of static noise but struggle significantly more with lower levels of dynamic noise. In contrast, few-shot prompted models appear more sensitive to even static noise. We conclude with a discussion of how our findings impact noise filtering best-practices, in particular emphasizing the importance of removing samples containing destructive dynamic noise with global errors.
\end{abstract}

\section{Introduction}

Step-by-step or chain of thought (\textbf{CoT}) prompting and its variants have been widely shown to improve large language model performance on numerous benchmarks \citep{Wei2022ChainOT, Chen2022ProgramOT, Gao2022PALPL, Wang2022SelfConsistencyIC, Zelikman2022STaRBR, Dohan2022LanguageMC}. Without additional fine-tuning, step-by-step reasoning capabilities only emerge in very large models (50B+ parameters) \citep{Wei2022ChainOT}. Recently, many works \citep{Ho2022LargeLM, Hsieh2023DistillingSO, Mukherjee2023OrcaPL, Gunasekar2023TextbooksAA, Xu2023WizardLMEL} attempt to distill these capabilities from large models into smaller models by sampling large numbers of CoT over a diverse set of tasks. Smaller models are then fine-tuned on the resulting synthetic dataset with the goal of improving task capabilities.

However CoT data, both internet-sourced and model-generated, is often noised with irrelevant or inconsistent steps \citep{LYU2023FaithfulCR, Turpin2023LanguageMD}. It is becoming increasingly common to heuristically filter out low-quality samples \citep{Schick2023ToolformerLM, Li2023SelfAlignmentWI} or hand-select high quality samples \citep{Zhou2023LIMALI} for best performance during both pretraining and fine-tuning. Yet even despite relatively poor training data quality, both large models and distilled smaller models are able to leverage CoT data with great improvement on downstream reasoning benchmarks. This prompts the research questions: 

\textbf{RQ1:} ``How does noise in CoT training data quantitatively affect model task performance?"

\textbf{RQ2:} ``How does noise quantitatively affect different phases of model training (e.g. pretraining, fine-tuning, prompting)?''
  
We start the study of these question in a controlled setting by working with \textit{algorithmic chains of thought} generated by tracing the execution of functions on lists of integers. Working in this setting allows for easy control over both the design of the CoT learned by the student and control over how much and what type of noise is introduced. Algorithmic tasks also make up a small percentage of model pre-training data, helping to remove confounding factors. One important component of a well-designed algorithmic CoT is its \textit{modularity} or decomposability into contextually independent sub-problems. This allows certain sub-problems in the chain to be optionally condensed or expanded depending on the desired level of detail. To allow for the easy contraction or expansion of sub-problems within an algorithmic CoT we design the \textit{traced int} (TInt) framework. TInt re-implements basic arithmetic on integers, allowing us to generate a programmatic trace while also controlling which parts of the full trace are made visible in CoT training data. Further, we can apply this framework to arbitrary functions on lists of integers, controlling trace visibility of the function at a highly granular level. 

In this work, we consider the tasks of arithmetic and median finding. We evaluated multiple CoT designs for the addition task and choose a design which maximizes performance while minimizing token length. Using this design we establish baselines in the noise-free case by studying the benefits of training with CoT over a \textit{direct} baseline with no CoT. Next, with baselines in place, we examine the effect of introducing noise. We classify each type of noise as either \textit{static}, if the noise is applied in post-processing without affecting future computation steps, or \textit{dynamic}, if the noise accumulates as the CoT is constructed, affecting future computation state values. Informally, static noise is meant to model reasoning traces with local errors which leave the high level structure of the computation unchanged. Dynamic noise models reasoning traces which contain missteps during computation which globally alter the form of the computation. For examples of statically and dynamically noised CoT, see Section \ref{sec:noisycot}.

From these experiments we find training with algorithmic CoT can tolerate surprisingly high levels of contamination without impacting performance when compared to learning the algorithmic tasks without any CoT. Remarkably, even when the entire dataset is contaminated with static noise, performance can remain unaffected. However, we find dynamic noise is more destructive, even at lower intensities. This provides evidence that during training models can robustly predict the next token in very noisy training data by attending to multiple prior token features (i.e. previous computations), allowing it to reliably predict the next token even if some of the features themselves are noised. Therefore, a model is naturally able to correct for errors in training data that only have a localized effect, allowing the model to look back far enough in-context to correct itself. However, performance significantly degrades when training data has errors with more global effects, as now \textbf{all} tokens downstream from an error are unreliable.\footnote{Code: \url{https://github.com/Dahoas/transformer_arithmetic
}}

In summary we make the following contributions: 

\begin{enumerate}
    \item The TInt framework for flexibly tracing and noising arbitrary python functions over integers.
    \item  A study of the performance of models fine-tuned and prompted on noiseless and noised algorithmic CoT. This reveals the robustness of CoT to all noise (compared to training with no CoT) and the extreme robustness of CoT to static noise.
    \item A discussion of the implications of our findings for training LLMs on natural language CoT.
\end{enumerate}

\section{Related Work}

\textbf{Chain of thought reasoning in LLMs}\quad Chain of thought prompting is a widely studied phenomena in language modeling \citep{Wei2022ChainOT, Gao2022PALPL, Wang2022SelfConsistencyIC, Zelikman2022STaRBR, Dohan2022LanguageMC, Shi2023LargeLM, Huang2022LargeLM, Chen2022ProgramOT} which significantly expands model problem solving capability. A large body of work has also been done studying why chain of thought prompting improves model performance \citep{Madaan2022TextAP, Wang2022PINTOFL, Wang2022SelfConsistencyIC, Min2022RethinkingTR, LYU2023FaithfulCR, Chung2022ScalingIL}. Surprisingly, \citet{Madaan2022TextAP} finds large models generate much irrelevant content in CoT and do not rely too heavily on intermediate reasoning for their final answer. A growing body of work also attempts to distill CoT traces from large teacher models into students \citep{Hsieh2023DistillingSO, Ho2022LargeLM}. Several papers also investigate fine-tuning models on algorithmic CoT generated as part of program's execution trace \citep{Nye2021ShowYW, Pi2022ReasoningLP} or arithmetic \citep{Nogueira2021InvestigatingTL, Qian2022LimitationsOL, Zhou2022TeachingAR, Liu2023GoatFL, lee2023recursion, Muffo2023EvaluatingTL}.

\textbf{Learning from noisy data in NLP}\quad A collection of work in NLP studies the robustness of models to noisy input data \citep{al-sharou-etal-2021-towards, Song2020LearningFN}  with many papers focusing on the imapct of noise on Neural Machine Translation systems \citep{Pruthi2019RealisticNT, Belinkov2017SyntheticAN, sperber-etal-2017-toward, anastasopoulos-etal-2019-neural}. These papers tend to focus on different aspects of noise than we consider such as grammatical errors, jargon, and mis-translations. Other works focus on learning specific tasks with labels (but no CoT) \citep{Wu2023NoisywikiHowAB, Hedderich2021AnalysingTN} which become especially relevant for crowd-sourced datasets \citep{Xiao2015LearningFM, Chen2020BeyondCA}.

Despite the above work on the impact of noisy training labels on learning NLP tasks, little work investigates how noise in CoT structured data impacts task learning performance. Yet understanding this effect is crucial, as CoT style training and inference account for the majority of modern LLMs' impressive reasoning capabilities. We initiate the quantitative study of noise in algorithmic CoT training data. 

\section{Methods: Generating Noisy Algorithmic Chain of Thought}
\label{sec:noisycot}

\begin{figure}
    \centering
    \includegraphics[scale=0.27]{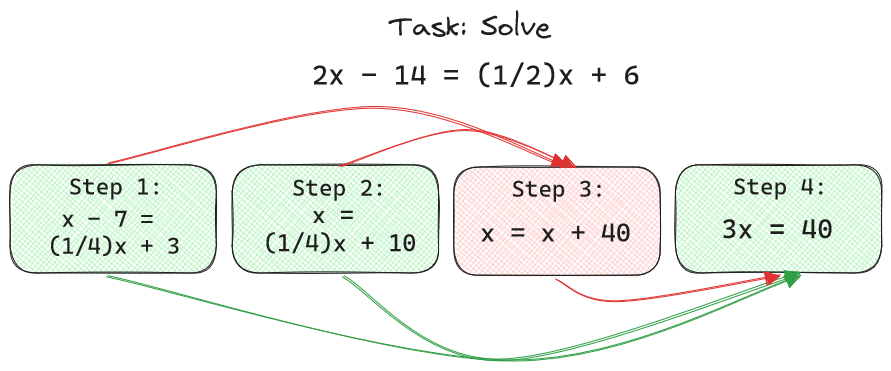}
    \caption{Example of an LLM preventing error propagation by attending to all prior steps. Naievely, the LLM might predict 0 = 40 after the incorrect Step 3. However, Steps 1 and 2 act as a mechanism to ``downvote" the bad influence of the incorrect Step 3 while simultaneously ``upvoting'' the correct Step 4 prediction.}
    \label{fig:noise-robustness}
\end{figure}

Algorithmic CoT trades off the ``difficulty" of predicting any one token in a computation by spreading the difficulty over many easier-to-predict tokens. From a computational perspective, this allows the model to perform more forward passes per problem, offloading computation done in the model weights to computation done in language. There are several important properties of interest benefiting from algorithmic CoT training including model performance, length generalization, task composability (some of which we benchmark in appendix Section \ref{sec:cot-design}).

Most interesting to us is the ability of CoT to improve robustness to training noise when solving a task involving complex reasoning. While perhaps initially surprising, we hypothesize that CoT provides a way for the model to ``check its work'' by attending to multiple prior steps (some of which may contain noise), distributing and decreasing reliance on any single step. This property would make the LLM particularly robust to static/local forms of noise in which the model can still find reliable prior steps to attend to. See Figure \ref{fig:noise-robustness} for an illustration of this property when solving a linear equation.

\subsection{The TInt Framework}
To design and test different types of algorithmic CoT we implement the traced int or \textit{TInt} framework. TInts behave as non-negative integers, with the added benefit that we can recover a programmable trace of any arithmetic or computation involving a TInt. This enables generation of diverse algorithmic traces for functions taking TInts as arguments.

We generate CoT examples from functions in three steps: 1. First we sample integer inputs from a chosen distribution. 2. We execute a function using the sampled inputs, only saving pieces of the trace marked as \textbf{visible} to the TIntegers. 3. We clean the accumulated trace, optionally removing code lines, inserting spaces between integers, and reformatting into more natural looking text. 

\begin{figure}
    \centering
    \includegraphics[scale=0.28]{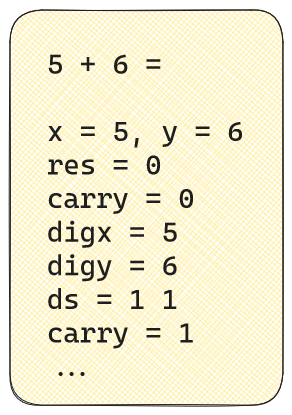}
    \caption{Example prefix of an algorithmic CoT for addition.}
    \label{fig:cot-example}
\end{figure}

\textbf{Step 1: Choosing a good integer sampling distribution}\quad We employ two methods of sampling integer arguments. The first, \textit{magnitude sampling}, is simply uniform random sampling between given lower and upper bounds $l$ and $u$. The second method, \textit{length sampling}, first uniformly samples the length $d$ of a number, then independently sample each digits where the first digit is non-zero. In practice length sampling bsignificantly improves length generalization, so we use this by default.

\textbf{Step 2: Algorithmic chain of thought trace design and accumulation}\quad For each set of sampled arguments we next trace their exeuction with with the target function. For each step in the code, we have available information on the state of variables as well as the line of code. This gives much flexibility in terms of what parts of the execution the model is allowed to see, allowing us to explore the trade-off between more and less detailed chain of thought. More detailed chain of thought makes it difficult to train the model on longer problems, but less detailed chain of thought requires the model to implicitly learn omitted computation. 

We found including lines of code used to generate traces harmed performance while taking up extra context. Our best performing CoT design outputs only the delta of variable values on each line, which is more context-efficient. For example, if in stage i, the state can be represented by $x = 5, y = 6$, and the line of code assigns $x = 7$, the print for that line is $x = 7$, with the value of y omitted.  

During this stage we also omit sub-routines in the TInt framework whose visibility has been explicitly set as invisible. For example, during multiplication we do not show the trace for addition as the resulting entire trace would be far too long for training. By default we make only very low-level operations, such as checking for equality, right-shifting, left-shifting, or appending to a list invisible.

\textbf{Step 3: Post-processing the accumulated trace}\quad The final step is to post-process the trace. An important step in cleaning inserts spaces between digits. As first noted by \citet{Nogueira2021InvestigatingTL}, this is important for models whose tokenizers do not tokenize digits individually. During this phase we also remove code-lines from the trace and add static noise (if present).

\subsection{Adding noise to algorithmic CoT}

The TInt framework allows us to inject noise into the generated algorithmic traces either during accumulation or post-processing. We refer to noise injected during computation time as \textit{dynamic} and noise injected during post-processing as \textit{static}. The two are quite distinct as a static noise error has only a local effect, whereas a dynamic noise error has a more global effect by affecting the entire subsequent trace. We call the fraction of samples containing noise in the dataset the \textit{dataset noise level} $n_d$ and the amount of noise in each individual sample the \textit{noise intensity}.

We experiment with two types of static noise: character level and line level. Character noise is injected by first setting a character noise intensity value $n_c$ and randomly flipping each digit in an affected algorithmic CoT with probability $n_c$. Line level noise operates similarly per sample with a paramter $n_l$ determining the probability each line in the trace gets deleted. Dynamic noise is introduced via the intensity parameter $n_{dl}$ which with probability $n_{dl}$ flips the digits of a newly initialized TInt during trace accumulation.

In addition, to modeling various types of local and global noise, we also believe these modeling choices reflect the type of noise often encountered in CoT in the wild. Character level noise corresponds to model minor hallucinations or missteps which are ignored later on. Line level noise is particularly common, as both model generated and internet-sourced CoT frequentyly is underspecified, containing (sometimes large) gaps in detail. Dynamic noise represents CoT which contains more global errors, resulting in solutions which are not even close to the correct answer.

\begin{figure}
    \centering
    \includegraphics[scale=0.55]{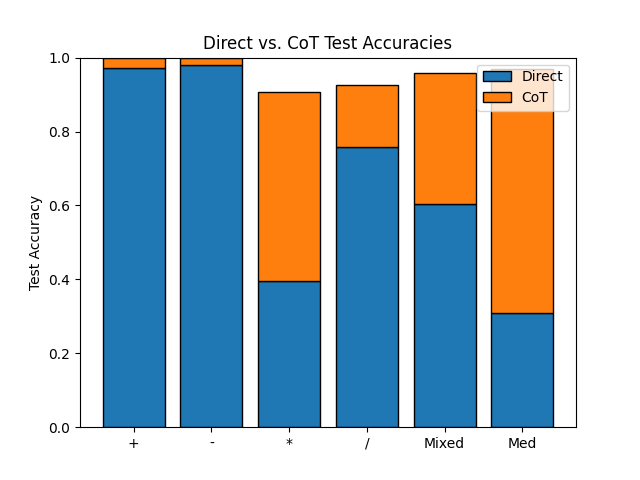}
    \caption{Test accuracy on up to 10-digit addition, 10-digit subtraction, 5-digit multiplication, 5-digit division, and up to 10 number median finding. Using Algorithmic CoT significantly improves performance on all tasks. Note: The equal mixture task trains and evaluations models on all four arithmetic operations. Note: Direct training is done for 100 epochs (10 more than for CoT).}
    \label{fig:noise-free-sft-accs}
\end{figure}

\begin{figure*}[ht]
    \centering
    \includegraphics[scale=0.5]{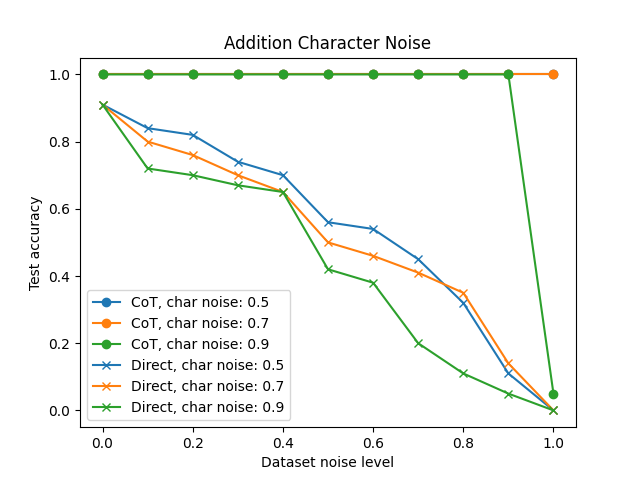}
    \includegraphics[scale=0.5]{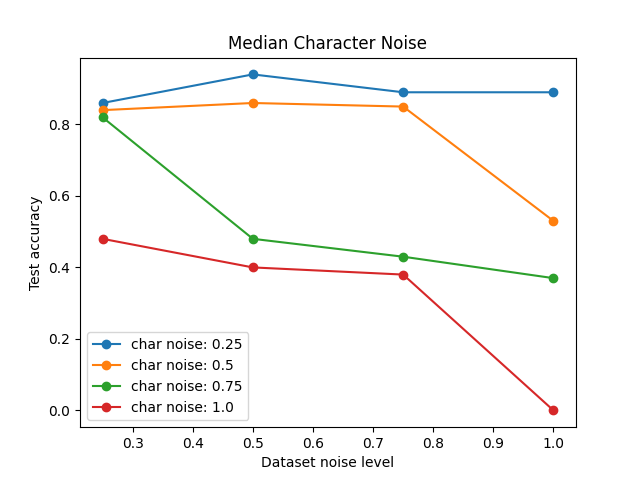}
   \caption{Plot of model test accuracy on len 1-10 addition and median vs. the dataset noise level (percent of samples with noise). Multiple levels of character noise intensity $n_c$ (percent of randomly flipped digits in a noised sample) are plotted. Algorithmic CoT retains perfect accuracy on all noise levels except when $n_d = 1.0, n_c = 0.9$. CoT refers to experiments with CoT training data while Direct refers to experiments with training data of only the answer. The median task also remains unaffected by noise when the character noise intensity is low ($< 0.5$) but appears more sensitive at high levels of dataset contamination and noise intensity.}
   \label{fig:prompt-static}
\end{figure*}

\section{Experiments}

For algorithmic tasks we choose arithmetic operations (addition, subtraction, multiplication, division) and the list median finding task. To implement algorithmic CoT for arithmetic we use ``grade school" algorithms. Addition and subtraction are treated as atomic when implementing multiplication and division to save context. The median algorithm used first selection sorts the array and then picks the middle element or average two middle elements. Experiments are run in two training regimes: via supervised fine-tuning and prompting (with prompting being considered a ``light'' form of fine-tuning \citep{Oswald2022TransformersLI}). 

\subsection{Fine-tuning on Noisy Chain of Thought}

We fine-tune Pythia-410m \citep{Biderman2023PythiaAS} with a context length of 2048. Note, as Pythia's tokenizer does not tokenize integers separately, a spaced representation is crucial for good performance. We train on 20,000 samples generated via length sampling up to length 10 for subtraction and addition, and up to length 5 digits for multiplication and division. Additionally we train on up to 10 five-digit integers for median finding. We use a batch size of 32 and lr = 1e-4 with 10 epochs. We test on 1000 samples drawn from the same distribution as the training data.

We start by first establishing fine-tuning performance baselines in the noise-free setting. To understand the performance benefits of CoT versus no CoT, we train models both with and without algorithmic CoT and report the results in Figure \ref{fig:noise-free-sft-accs}. CoT significantly improves performance across the board and particularly for harder to learn functions like multiplication and median finding. We conduct a more thorough study of the impact of CoT on task performance, sample-efficiency, and length generalization in appendix section \ref{sec:noise-free}, finding significant improvements when compared to direct prediction.

With baselines established in the noiseless case, we begin applying noise to the target. We start with static noise, applying digit-level noise with parameter $n_c$ controlling the probability of a digit in the target getting randomly flipped. As a warmup, we train models as before on length sampled addition from 1-10 digits both directly and with CoT for median finding. We also vary the dataset noise level $n_d$ from $0$ to $1$ which controls the percent of samples containing any amount of noise. We do this both for CoT solution traces as well as direct solution traces. Note, we never noise the input arguments, only the intended output. Results for addition and median finding tasks are reported in Fig. \ref{fig:prompt-static}.

\begin{figure*}[ht]
    \centering
    \includegraphics[scale=0.5]{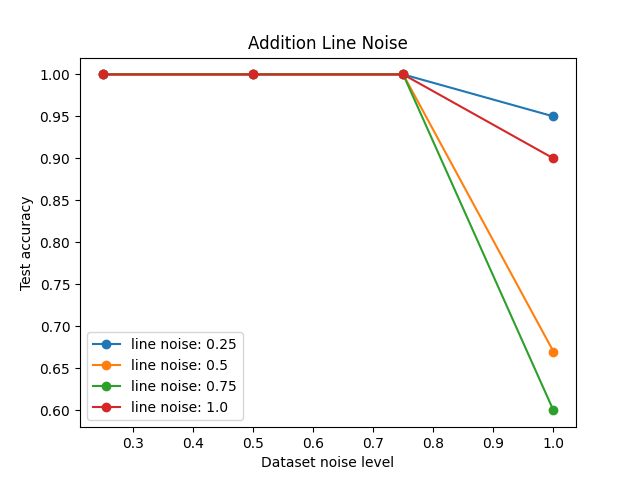}
    \includegraphics[scale=0.5]{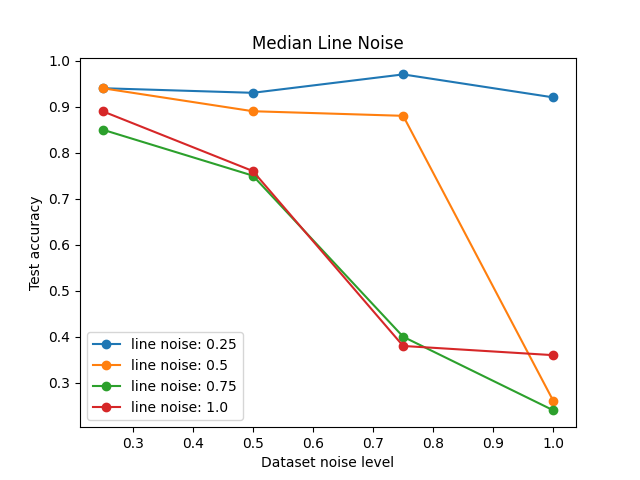} 
   \caption{Plot of model test accuracy on len 1-10 addition and median vs. the dataset noise level (percent of samples with noise). Multiple levels of line noise intensity $n_l$ (percent of randomly deleted lines in a noised sample) are plotted. As with character level noise, both tasks are robust to lower levels of line noise intensity and datase noise level, with addition being extermely robust at even higher levels. }
   \label{fig:line-noise}
\end{figure*}

\textbf{Character-level static noise}\quad The results for addition demonstrate algorithmic CoT is extremely robust to the dataset noise level, with the model's performance unaffected until the entire dataset is contaminated. Remarkably, the model perfectly learns the algorithm even with 70\% of all digits corrupted. Only when 90\% of all digits are corrupted in all samples does the model fail to learn anything. In contrast, training directly on solutions is much more sensitive to the dataset noise levels. Performance decreases linearly to 0 as dataset noise level goes to 1, with character noise level only weakly affecting decay rate. Taken together, these results on addition suggest \textbf{CoT significantly improves robustness to character level static noise} during training. 

While similar overall, the median task appears more sensitive to character noise level. A model trained with 25\% character noise remains largely unaffected at all dataset noise levels. But as character noise increases, we see performance start to drop near $n_d = 1$, with $n_c = 1, n_d=1$ leading to $0$ accuracy. This is expected as the median CoT requires the model to learn more complex primitives like division by two. Reliably learning these primitives is much harder in the noisy setting. When we examine samples on which CoT fails to correctly compute the median, we see it is usually on even length lists where the mean is incorrectly computed. This explains the drop to about 0.5 accuracy on higher noise levels. This means that unless the noise levels are both quite high, the model largely retains the ability to `pick' and copy the middle number.

\textbf{Line-level static noise}\quad We also experiment with applying a second type of static noise, line-level noise, which deletes lines from the algorithmic CoT. To compare with our results on character level static noise we run experiments with both dataset noise levels and line noise levels (the probability of deleting any given line) ranging from $0.25-1$. Results on addition and median are reported in Fig. \ref{fig:line-noise}.

As observed with character level noise, \textbf{addition CoT remains largely unaffected until the dataset is fully contaminated}. Then, when $n_d = 1$, performance dramatically drops with the lowest reached by line noise intensity $n_l = 0.75$. However, we can recover accuracy up to $0.91$ when line noise intensity is $1$ i.e. all intermediate lines are deleted. This is not surprising, as in this case we removed the entire CoT and directly predict the answer. This hints that the most destructive line noise intensities delete most but not all intermediate lines, creating wide variance of chain of thought form in the data. Indeed, we reach our lowest accuracy of 0.6 when $n_d = 1$ and $n_l = 0.75$.

As with character noise, performance on median finding exhibits a similar trend with more sensitivity to the dataset noise level when the line noise intensity is higher. Interestingly, line noise intensity $n_l = 0.5$ exhibits good performance until the entire dataset is noised at which point the resulting model performs the worst overall. This reinforces the observation that lower levels of noise intensity can be tolerated, as the model is able to recover the entire form of the algorithm across multiple training samples.

\begin{figure}[ht]
    \centering
    \includegraphics[scale=0.55]{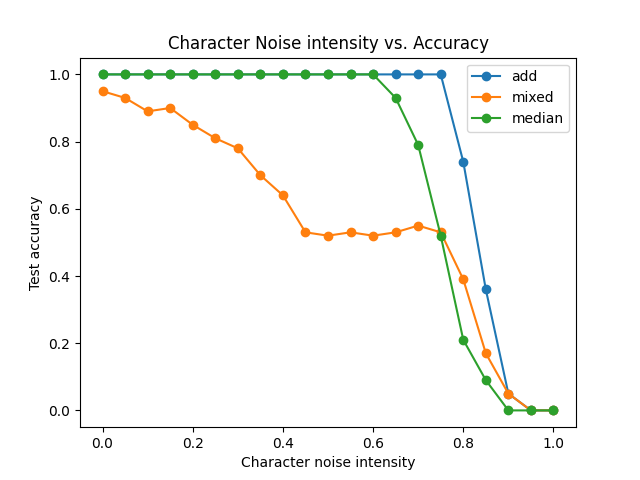}
   \caption{Plot of character noise intensity $n_c$ against CoT model performance on addition, mixed arithmetic, and median finding tasks.}
   \label{fig:fine-static-noise-accs}
\end{figure}

\textbf{Learning algorithmic CoT with no clean samples}\quad Our findings above indicate fine-tuning the transformer on a statically noised CoT dataset remains extremely robust until both the dataset noise level and the static noise level are too high ($\approx 1$). Due to this fact, we now focus on the case $n_d = 1$, to more precisely understand how the static noise level affects the performance decay rate. We now also evaluate the mixed arithmetic task, which requires the model to learn $+,-,*,/$ in equal distributions. Results can be found in Fig. \ref{fig:fine-static-noise-accs}. 

When holding the dataset noise level fixed at $n_d = 1$ and varying the character noise level $n_c$ from $0$ to $1$ we see a big difference in behavior between addition and the more complex mixed arithmetic and median finding tasks. The addition task appears unaffacted by $n_c < 0.75$ but swiftly drops to $0$ after $n_c \geq 0.75$. This phase transition is likely due to the model simply not having any clean sub-computations to reliably attend to, making correct prediction impossible.

We find that for the mixed arithmetic task, character level noise is more destructive than for addition. The failure cases indicate this is primarily due to atomic sub-routines like addition and subtraction being miscomputed in multiplication and division traces. This suggests character level noise is more harmful to CoT containing underspecified intermediate results from complex sub-routines. In comparison the addition task does all computation explicitly in the CoT, preventing collapse until higher character noise intensities.

We also apply line level static noise with $n_d = 1$ (See Appendix \ref{sec:additional-noisy-fine-tune}). These experiments confirm earlier observations that line noise is most destructive around $n_l = 0.7$, resulting in a decreasing then increasing relationship between line noise level and test accuracy.

\begin{figure}[ht]
    \centering
    \includegraphics[scale=0.55]{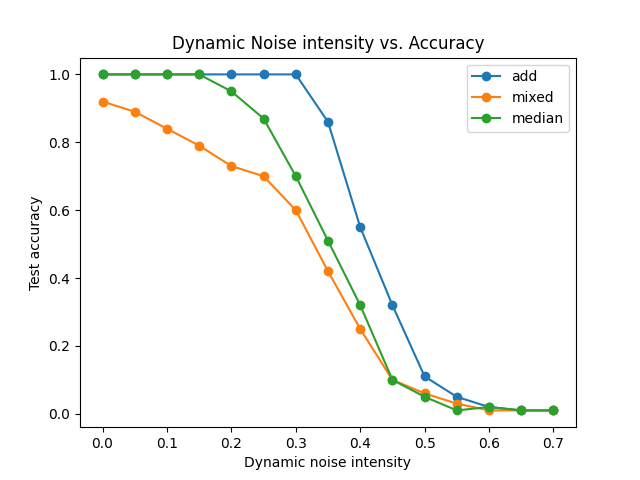}
   \caption{Plot of dynamic noise intensity against model test accuracy on algorithmic CoT addition, mixed arithmetic, and median finding. Dynamic noise is more destructive than static noise for test performance.}
   \label{fig:dynamic-noise-accs}
\end{figure}

\textbf{Adding dynamic noise}\quad Lastly, we experiment with dynamic noise. This noise is more destructive than static noise, as errors made earlier in a trace propagate. Since we have identified algorithmic CoT as being robust to low levels of dataset noise, we focus on the case $n_d = 1$ and vary the dynamic noise level $n_{dl}$. Results are reported in Fig. \ref{fig:dynamic-noise-accs}.

Compared to static noise, we find dynamic noise to be more destructive, with dynamic noise intensities above $n_{dl} = 0.5$ completely destroying performance on all tasks. In contrast, addition maintained perfect performance with static noise until $n_c, n_l \geq 0.8$. We believe the more global property of this noise--that it affects more tokens in the data--makes it difficult for the model to ignore this noise as well as it does the static noise. As result even with low dynamic noise, the model fails to learn the correct algorithmic CoT. Additionally the CoT examples injected with dynamic noise can generate significantly longer traces. This impacts performance, making it difficult for the model to choose the noise-free CoT sub-computations, which it could do in the static case. 

Dynamic noise is even more destructive for CoT with complex sub-routines like division by two in the median finding task. Above a low noise intensity ($n_{dl} = 0.3$) the model is unable to reliably learn how to correctly compute division by two or even when to use division. Similar problems occur for multiplication and division in the mixed arithmetic task, which themselves call addition and subtraction as atomic sub-routines. 

\begin{figure*}[t]
    \centering
        \includegraphics[scale=0.33]{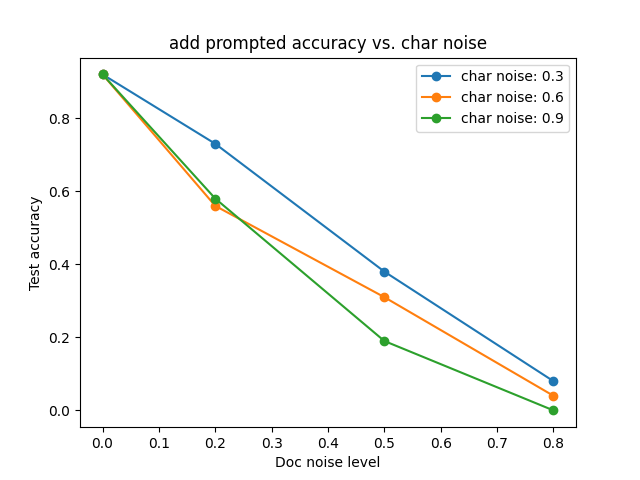}
        \includegraphics[scale=0.33]{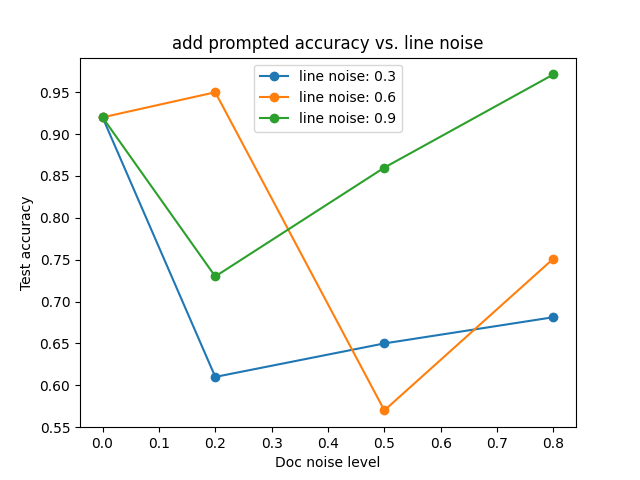}
        \includegraphics[scale=0.33]{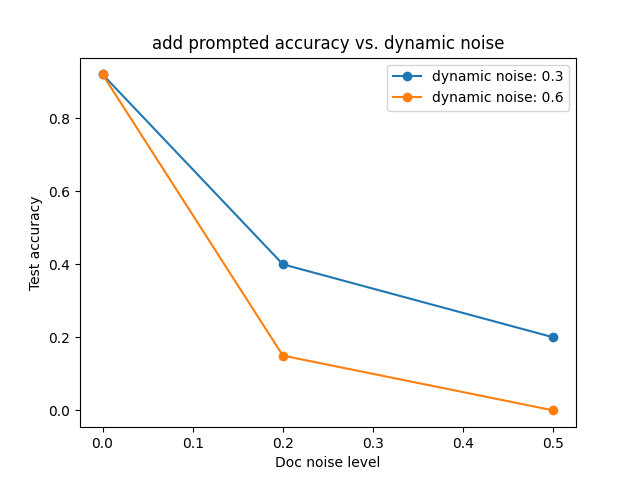}
    \hfill
        \centering
        \includegraphics[scale=0.33]{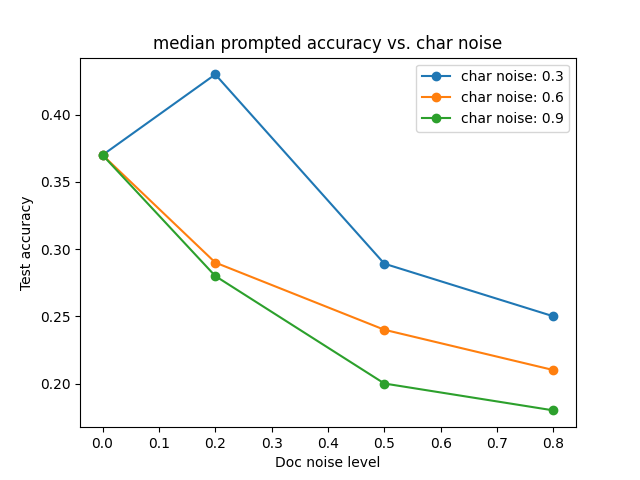}
        \includegraphics[scale=0.33]{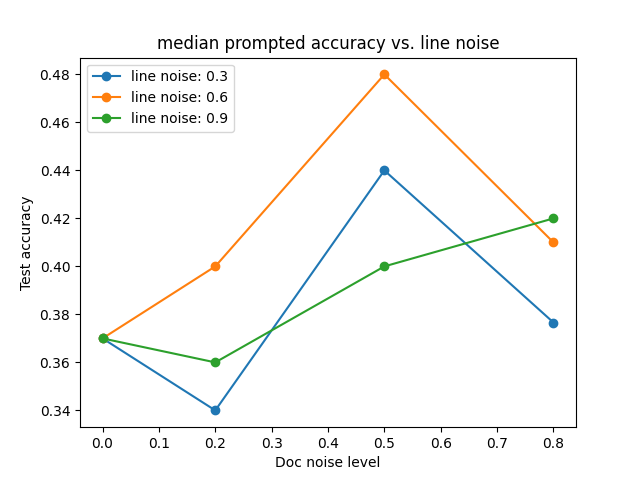}
        \includegraphics[scale=0.33]{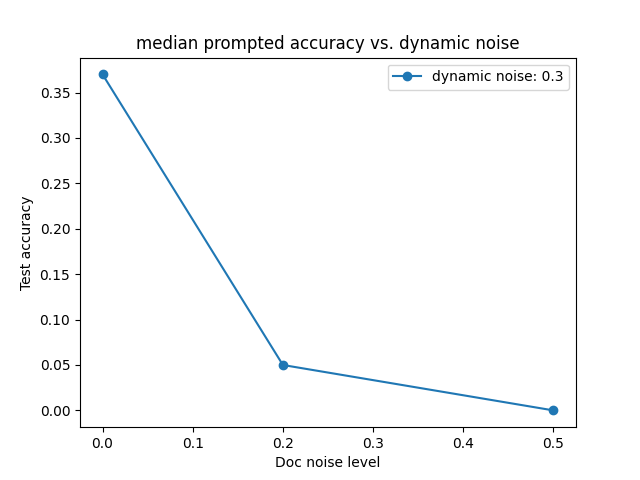}
    \caption{Plots of model test accuracy versus document noise intensity for various character noise levels on addition and median tasks. Note: ``doc-noise-level'' refers to dataset noise level, i.e. the percentage of demos in the prompt that have been noised.}
    \label{fig:prompt-noise}
\end{figure*}

\textbf{Discussion}\quad We believe the robustness CoT to high levels of static noise is attributable to two main factors. First, CoT training is known to be \textbf{highly sample efficient}, requiring far fewer samples than the direct analogue to learn the desired algorithm. This likely helps the model learn the target algorithm even when few noise free samples are available. Yet this is not a full explanation as even when the dataset noise level $n_d = 1$, models still are able to get good performance. We further hypothesize \textbf{LLMs are able to recover from local errors by attending to multiple prior steps} (some of which may contain noise), distributing and decreasing reliance on any single step. See Figure \ref{fig:noise-robustness} for an example of this property. The more harmful effect of dynamic noise further supports this hypothesis, dynamic noise has a global effect, corrupting all steps after an error. This makes it much more difficult for the model to look back in-context far enough to find correct steps it can use to help verify its next prediction.

\subsection{Prompting with Noisy Chain of Thought}

With our study of the impact of noise during fine-tuning done we now focus on the impact of noise during prompting. To do so we evaluate the performance of SOTA models (\texttt{gpt-4}, \texttt{gpt-3.5-turbo}) when prompted with 10-digit addition, 5-digit multiplication, and median finding on 5, 5-digit numbers. We generate prompting and testing datasets as before by sampling, based on length, integer arguments and then computing an algorithmic CoT using the TInt framework. 100 training prompts and 100 testing questions are sampled per task. We first establish baselines for how well models are able to complete the task given algorithmic CoT examples in context. Table \ref{tab:prompted-noise-free} reports the results of prompting models with both three and six in-context demonstrations. The specific prompt we use can be found in the appendix.

We find gpt-4 does a good job following the given algorithm for addition, but struggles with following the median and multiplication algorithms. There is a noticeable improvement from 3-shot to 6-shot, so by default we give a 6-shot promopt for our noisy experiments. gpt-3.5-turbo is unable to solve the tasks with the desired algorithmic framework, so we do not bother evaluating it in the noisy case. With baselines established, we can now study the impact of noise when prompting gpt-4 algorithmically.

\begin{table}[h]
    \centering
    \begin{tabular}{lcccc}
     & \multicolumn{2}{c}{\texttt{gpt-4}} & \multicolumn{2}{c}{\texttt{gpt-3.5-turbo}} \\ 
     \cmidrule(lr){2-5} & 3-shot & 6-shot & 3-shot & 6-shot \\
  \toprule
  Addition   & 0.86 & 0.92 & 0.0 & 0.0\\
  Median     & 0.33 & 0.37 & 0.0 & 0.0\\
  Multiplication  & 0.24  & .27 & 0.0 & 0.0\\
  \bottomrule
\end{tabular}
    \caption{Prompted accuracy of \texttt{gpt-4-1106-preview} and \texttt{gpt-3.5-turbo-1106} on algorithmic CoT following. gpt-4 does well on addition following the algorithm but performs much worse on median finding and multiplication. gpt-3.5 cannot follow the CoT at all.}
    \label{tab:prompted-noise-free}
\end{table}

\textbf{Few-shot prompting with noisy CoT}\quad We now apply noise to the in-context demonstrations, corrupting each one with probability equal to the dataset noise level $n_d$.  As above, we prompt gpt-4 with six examples. Figure \ref{fig:prompt-noise} contains plots of of gpt-4 test performance for addition and median tasks. We include results for multiplication in the appendix.




In comparison with fine-tuning, the prompted model seems far more sensitive to both the noise intensity and dataset noise level. Character noise on addition linearly decays to 0 with dataset noise level. Surprisingly, this is similar to the performance of the noisy fine-tuned model without any CoT. The median task accuracy decreases to 0.2 (equivalent to random guessing). The effect of static line noise is a bit more surprising: performance initially decreases, then starts increasing as dataset level noise increases. Recall, at 100\% dataset noise and 100\% line noise we recover the direct task. On addition gpt-4 exhibits impressive capabilities zero-shot. This explains the good performance for $n_l = 0.9$. However, we expect model performance to suffer with $n_l < 0.9$ as in the fine-tuning case. Yet this is not the case, indicating somehow the prompted model is more robust to line level noise. Finally, dynamic noise completely destroys model performance on both tasks, even at small dataset noise levels. In comparison to fine-tuning, prompting seems even more sensitive to this type dynamic noise.

\textbf{Discusssion}\quad We note the prompted model's sensitivity to noise is somewhat in contradiction with other works which find prompting models with incorrect natural language CoT does not significantly impact performance \citep{Madaan2022TextAP}. One key difference here is the usage of algorithmic CoT which specifies an algorithm likely not well-represented in the training data. This forces the prompted model to learn the intended algorithm strictly from the in-context examples on certain types of tasks. This is not necessarily the case with natural language CoT prompting. The result is a higher sensitivity to all types of noise. We also mention our prompting results agree with what has been observed in other work prompting LLMs to do algorithmic CoT \citep{Zhou2022TeachingAR}. In particular, it was previously observed LLMs are somewhat robust to the low-levels of ``untargeted'' noise but are more susceptible to ``targeted'' noise designed to confuse the model.

\section{Conclusion and Broader Impact}

We studied the robustness of LLMs to various types of noise when training on alorithmic CoT structured data. We found models fine-tuned on datasets containing static, local noise are remarkably robust to both the dataset noise level and the noise intensity. In contrast dynamic noise, which results in more global errors, is much more destructive. Prompted models appear to exhibit the same trends, but with a higher overall noise sensitivity. We then discussed how LLMs remain so robust to static noise, citing both the sample efficiency of CoT and the ability of the LLM to attend to multiple prior steps to self-correct instead of over-relying on the most recent noisy computation.

Our findings suggest it is critical to filter out documents containing large amounts of dynamic, global noise during both pretraining and fine-tuning. However, it appears less necessary to filter out CoT data containing low or moderate amounts of static noise. Such data is likely even helpful if the noise intensity is not too high. Our findings also partially help explain the successes seen more recently in distilling CoT data to smaller models. Generally LLMs seem to have an understanding of the high-level solution but struggle with getting all the details right. This type of CoT data, often generated by a larger model, will likely have large amounts of static noise but lower amounts of dynamic noise.

\textbf{Limitations and Future Work}\quad For future work we intend to study the impact of scale, as we expect larger models to be more robust to all types of noise during training. Also, our study focuses exclusively on training models via fine-tuning and prompting. It is also of interest to see if there are similar trends for pretraining and in particular how the  distribution of pretraining data affects noise sensitivity downstream. 

\bibliography{main}
\bibliographystyle{main}

\appendix

\begin{figure*}[ht]
    \centering
    \includegraphics[scale=0.5]{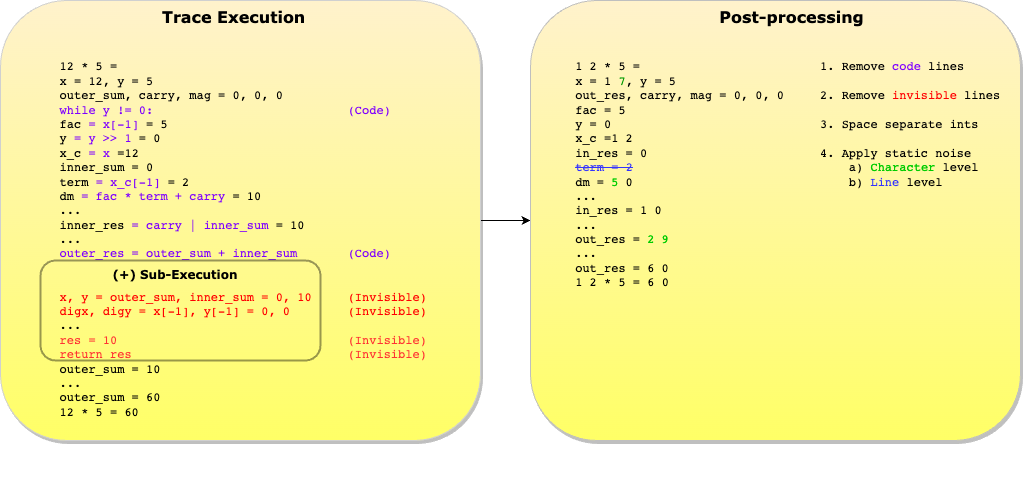}
   \caption{The traced multiplication of $12 * 5$ in two steps. First, the execution trace consists of all code lines and variable updates. Sub-calls made to addition are configured to be \textit{invisible}, removed at trace execution time. Second, post-processing is configured to remove lines of code, leaving only visible variable updates. Spaces are then inserted in-between integers. Finally, static noise gets applied, either by randomly flipping integers or deleting lines.}
   \label{fig:tint}
\end{figure*}

See Fig. \ref{fig:tint} for more detail on the post-processing of the computation trace. 

\section{Baselines for Noise-Free Algorithmic Chain of Thought}\label{sec:noise-free}

As a baseline, we define the tasks of 10-digit addition, 10-digit subtraction, 5-digit multiplication, and 5-digit division. We also define the mixed task where 25\% of samples are allocated to each of the four tasks. To demonstrate the flexibility of our TInt framework we additionally include a median finding task for lists of integers up to length 10 with at most five-digit numbers.

\textbf{A case study in addition}\quad We begin by thoroughly evaluating the benefits of algorithmic CoT with addition as the exemplar. First we observe performance gains made by simply space separating integers for tokenization. Results are reported in Table \ref{table:nospace-acc}. Interestingly CoT gives a big improvement even when digits are not tokenized independently, likely as the CoT for addition expliclty isolates the digits at each decimal place. This confirms space separation also helps smaller decoder only models like Pythia.

\begin{table}[ht]
\caption{Accuracy on 5 digit addition.}
\label{table:nospace-acc}
\begin{center}
\begin{tabular}{ p{0.25\linewidth} c c} 
 \hline
   & no space & space \\ 
 \hline
 Direct & 0.25 & 0.94 \\
 \hline
 A-CoT & 0.95 & 1 \\
 \hline
\end{tabular}
\end{center}
\end{table}

\begin{figure}[ht]
    \centering
    \includegraphics[scale=0.55]{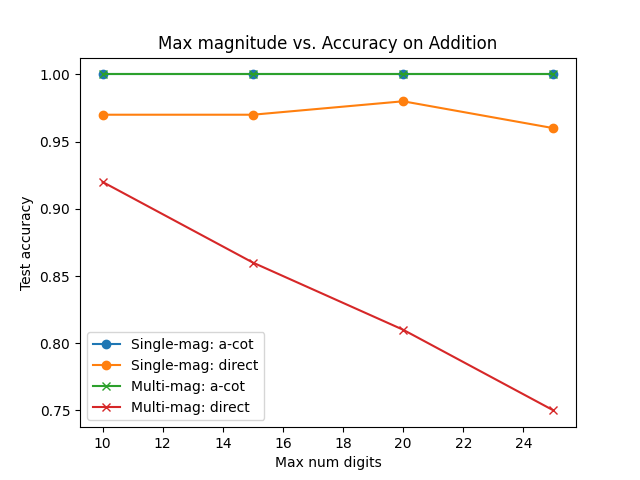}
   \caption{Plotting the maximum number of digits present in training against model accuracy. Single-mag models are trained and evaluated on a single digit length. Multi-mag models are trained and evaluated on a uniform distribution up to the maximum digit length. Direct models have no CoT while a-cot models train with CoT.}
   \label{fig:len-accs}
\end{figure}

Next with a fixed space-separated tokenization scheme we investigate how the sampled integer distribution affects model performance. We sample in two ways by either sampling integers of a single fixed length or sampling integers uniformly up to a maximum length as described in Section \ref{sec:noisycot}. Results are reported in Fig. \ref{fig:len-accs}.

Surprisingly we find our relatively small 410m parameter model does well when adding numbers of a fixed length. However, it struggles to generalize to the variable-length addition task, suggesting a tendency to overfit to a fixed length. In comparison algorithmic CoT training generalizes perfectly across all lengths in the training distribution. We remark algorithmic CoT training is highly data-efficient, requiring only a single epoch of training. In addition it is highly sample efficient, requiring only 1000 samples across all lengths to perfectly generalize. This suggests the CoT is forcing the model to learn the underlying algorithm exactly, instead of learning an approximate algorithm with unknown edge-cases. To check this hypothesis we train two models on the direct (no CoT) 50-digit addition problem. The model trained on only exactly 50-digit samples achieves near perfect accuracy, while the model trained on numbers sampled uniformly in length achieves only 0.36 accuracy.

These results also naturally lead to the question of how well algorithmic CoT generalizes to lengths outside the training data. To answer this we train models on data sampled with integers uniformly across lengths both directly and with algorithmic CoT. Results are reported in Table \ref{table:length-gen-acc} where we find algorithmic CoT significantly improves length generalization compared to direct training which exhibits very poor generalization. We also note training on fixed-length integers results in very poor generalization even for algorithmic CoT. 

However, even with algorithmic CoT the model fails to generalize beyond training length arbitrarily well. Curiously this failure is most often due to the model failing to copy inputs into the algorithm format correctly, as opposed to incorrectly executing the algorithm. For example, it might be the case 1234+9852 gets copied as x=123, y=985 in the next line, completely missing the least significant digits.

\begin{table}[ht]
\caption{Length generalization on addition. 1-L is sampled by length uniformly up to L. L as a constant fixes the length of testing samples.}
\label{table:length-gen-acc}
\begin{center}
\begin{tabular}{ p{0.25\linewidth} c c c c c c} 
 \hline
   & 1-15 & 15 &  1-20 & 20 & 1-25 & 25 \\ 
 \hline
 Direct 1-10 & 0.63 & 0 & 0.44 & 0 & 0.29 & 0 \\
 \hline
 Direct 1-15 & 0.84 & 0.2 & 0.59 & 0 & 0.37 & 0 \\
 \hline
 Direct 1-20 & 0.89 & 0.77 & 0.75 & 0.12 & 0.54 & 0 \\
 \hline
 A-CoT 1-10 & 0.84 & 0.2 & 0.63 & 0.09 & 0.45 & 0 \\
 \hline
 A-CoT 1-15 & 1 & 1 & 0.97 & .88 & 0.79 & 0.24 \\
 \hline
 A-CoT 1-20 & 1 & 1 & 1 & 1 & 0.99 & 0.85 \\
 \hline
\end{tabular}
\end{center}
\end{table}

\begin{table}[ht]
\caption{Test Accuracy on up to 10-digit addition, 10-digit subtraction, 5-digit multiplication, 5-digit division, and up to 10 number median finding.}
\label{table:baseline}
\begin{center}
\begin{tabular}{l c c}
     & No CoT & A-CoT \\ \hline
  10-digit addition       & .972 & 1 \\  \hline
  10-digit subtraction    & .979 & 1 \\  \hline
  5-digit multiplication  & .396 & .907 \\  \hline
  5-digit division        & .757 &  0.925 \\ \hline
  Equal mixture           & .604 & .958 \\ \hline
  10-median               & .306 & .97 \\ \hline
\end{tabular}
\end{center}
\end{table}

\textbf{Multi-task training on arithmetic}\quad Now having a better understanding of the benefits of algorithmic CoT, we train on all arithmetic tasks. Recall we train with length-sampled integers up to 10 digits on addition and subtraction, and up to five digits on multiplication and division. The same model is trained simultaneously on all tasks in a multi-task setting. Results are in Table \ref{table:baseline}. Even for the direct model addition and subtraction remain relatively easy to learn, with addition increasing in accuracy by 0.05 from the single-task case. However the direct model struggles to learn division and multiplication in particular, often failing completely for inputs with more than two digits. This in particular highlights the capabilities of algorithmic CoT which significantly improves performance on multiplication and division. We note however we do not achieve perfect performance on multiplication as we treat addition as atomic, requiring the model to learn up to five digit addition with no CoT.

We similarly compare the performance of learning the median function on list of integers up to length 10 with integer lengths ranging from 1-5. The direct model struggles here, whereas training with algorithmic CoT which first sorts the list and then extracts the median solves this easily. Note we do not treat addition as atomic, allowing the model to exactly learn the algorithm. Interestingly we observe direct models are very capable of learning both sorting and addition tasks independently. However, without CoT the model is unable to compose the two routines to solve median.

We also investigate the sample efficiency of CoT, plotting the number of samples against test accuracy in Figure \ref{fig:sample-efficient}. This confirms CoT as being incredibly sample efficient.

\begin{figure}
    \centering
    \includegraphics[scale=0.5]{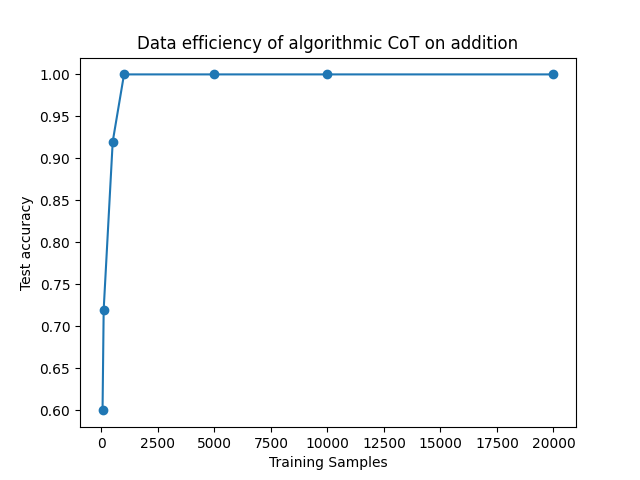}
    \caption{Sample efficiency of CoT training.}
    \label{fig:sample-efficient.}
\end{figure}

\section{Additional Results for Noisy Fine-tuning}
\label{sec:additional-noisy-fine-tune}

\begin{figure}[ht]
    \centering
    \includegraphics[scale=0.55]{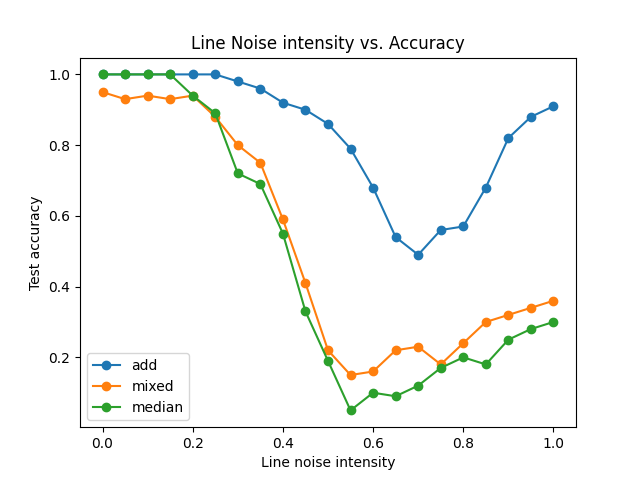}
   \caption{Plot of model test accuracy vs. line noise intensity with dataset noise level $n_d = 1$ (a fully noised dataset). }
   \label{fig:line-noise-accs}
\end{figure}

We apply line level noise with parameter $n_l$ as a second type of static noise. Recall a sample with line level noise deletes a line with probability $n_l$ for every line. Similarly to the character level case, algorithmic CoT is again robust to noise until the entire dataset is contaminated. Results are reported in Fig. \ref{fig:line-noise-accs}.

The experiments suggest line deletion is more destructive for all tasks, in particular median finding. This is not surprising as the median finding algorithm relies on a sorting of the input which relies more heavily on multiple control flows versus a task like addition which has a single control flow. However at $n_l = 0.5$ line noise destructively impacts all model performance, causing models to do far worse than even training with no algorithmic CoT. Interestingly the model learns to ignore irrelevant and unhelpful lines at extremely high levels of line deletion. As we approach deletion of the entire algorithmic CoT the model recovers its original direct task performance level. This suggests the most destructive type of line noise is around the 50\% deletion rate where the CoT generated have the highest diversity, meaning the model cannot properly recover the full algorithm.

\section{Additional Results for Noisy Prompting}
\label{sec:additional-noisy-prompt}

As a baseline, we evaluate performance of ten-digit addition with three-shot and six-shot prompting. In these experiments, we magnitude-sample three and six example traces, respectively, for each of 100 magnitude-sampled test cases. 

\begin{center}
\begin{tabular}{l c}
     & 10-digit addition accuracy \\ \hline
  three-shot   & .86 \\  \hline
  six-shot     & .92 \\  \hline
\end{tabular}
\end{center}

Notably, and as expected, there is slight performance increase when increasing the included samples in the prompt. For the rest of the experiments, we use six-shot prompting. 

As another baseline, we evaluate performance of ten-digit addition of two numbers, three-digit median finding on up to five numbers, and five-digit multiplication of two numbers:

\begin{center}
\begin{tabular}{l c}
     & accuracy \\ \hline
  addition   & .92\\  \hline
  median finding     & .37 \\  \hline
  multiplication     & .27 \\  \hline
\end{tabular}
\end{center}

From these results, we conclude median-finding and multiplication are operations more complex than addition, which therefore result in lower accuracy. Interestingly, the multiplication result suggests that the moderate line noise level creates the worst performance out of the three line noise levels tested. 

\begin{figure*}[t]
    \centering
    \begin{subfigure}[t]{0.5\textwidth}
        \centering
        \includegraphics[scale=0.5]{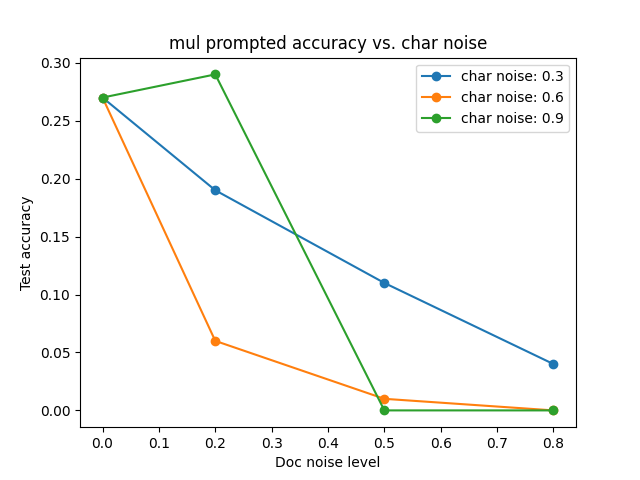}
    \end{subfigure}%
    \begin{subfigure}[t]{0.5\textwidth}
        \centering
        \includegraphics[scale=0.5]{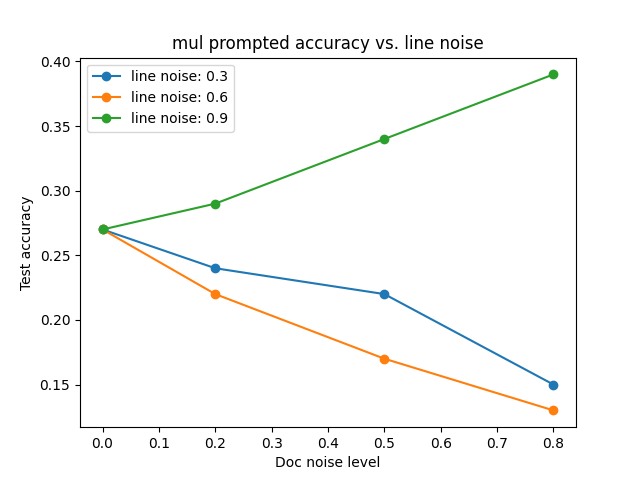}
    \end{subfigure}%
    \caption{Plots of model test accuracy versus document noise intensity for various character noise levels on the multiplication task. Overall, performance is similar to the median task, with more sensitivity the character level noise and line level noise.}
    \label{prompt_static}
\end{figure*}

We report noisy prompting results for multiplication in Figure \ref{prompt_static}.

\section{CoT Design}
\label{sec:cot-design}

In this work we explored various CoT design choices. We found it easiest to experiment with code-like CoT to control visibility of details and injection of noise. Note that some variants include code and natural language. After evaluation, we concluded representing algorithmic CoT with variable state assignment but without other lines of code to be the most effective. Experiments suggested including code lines at best are irrelevant and at worst detrimental to performance. Removing these lines has the added benefit of saving context. We found a similar effect with removing natural language text.

We also found well-designed algorithmic CoT often exhibits good \textit{locality} by propagating information needed later in the computation through the trace. In practice, this motivates design choices like accumulating partial final answers in a \texttt{res} variable and opting to iteratively remove off the least significant digits of input TInts $x$ and $y$ instead of attempting to index the appropriate decimal place.

Ideally all ``sufficiently complex" sub-routines, e.g. addition sub-steps in multiplication, are themselves parameterized with CoT. Otherwise the model must learn to perform them in a single forward pass. However due to context length requirements it is not always feasible to include all sub-routines.


\begin{small}
\begin{lstlisting}[caption=Traced Int Sample]
2 4 * 8
x = 2 4 , y = 8
out_res = 0
carry = 0
mag = 0
fac = 8
y = 0
x_c = 2 4
in_res = 0
term = 4
x_c = 2
dm = 3 2
in_res = 2
carry = 3
term = 2
x_c = 0
dm = 1 6
dm = 1 9
in_res = 9 2
carry = 1
in_res = 1 9 2
carry = 0
in_res = 1 9 2
mag = 0
out_res = 1 9 2
1 9 2
\end{lstlisting}
\end{small}

Additionally, many complex algorithms can be decomposed into simpler routines. For example, multiplication naturally decomposes to simpler operations like addition. We can use this to simplify traces for complex operations by replacing addition sub-traces with a call to the call($\cdot + \cdot$) api where the model calls itself as a sub-routine with input $\cdot + \cdot$ similar to learned tool usage \citep{Zhou2022TeachingAR, Schick2023ToolformerLM}. This would allow us to study the model's ability to learn problem decompositions and has the added benefit of dramatically shrinking trace sizes for more complex operations, which is important for models of smaller context length.

Previous work \citep{Nye2021ShowYW, Muffo2023EvaluatingTL, Qian2022LimitationsOL} utilize varying algorithmic chain of thought designs. \citet{Nye2021ShowYW} print out full execution trace data including line numbers and variable states while others \citet{Qian2022LimitationsOL, Zhou2022TeachingAR} design their chains in natural language. The TInt framework is designed for maximum flexibility over execution visibility. This allows us to explore design decisions for CoT involving how much detail to include in the execution trace. Our best performing CoT design outputs only the delta of variable values on each line, which is also more context-efficient. For example, if in stage i, the state can be represented by $x = 5, y = 6$, and the line of code assigns $x = 7$, the print for that line is $x = 7$, with the value of y omitted.

\begin{table*}[ht]
\label{table:CoT-examples}
\begin{center}
\begin{tabular}{|p{0.45\linewidth}|p{0.45\linewidth}|} 
 \hline
 \parbox[t]{35mm}{5 + 6\\
x = 5 , y = 6\\
res = 0\\
carry = 0\\
digx = 5\\
digy = 6\\
x = 0\\
y = 0\\
ds = 1 1\\
res = 1\\
carry = 1\\
res = 1 1 \\
1 1}
 & 
 \parbox[t]{35mm}{5+6=Let's add the 0 digits. 0 digit of 5=5. 0 digit of 6=6. Then, summed with carry, we have 11. 0 digit is 1. The carry is now 1. The result so far is 1. Adding the carry gives 11. Answer: 11}
 \\ \\
 \hline
 \parbox[t]{35mm}{5 + 6\\
...\\
digx = x[0]\\
digx = 5\\
digy = y[0]\\
digy = 6\\
ds = digx + digy\\
ds = 1 1\\
...\\
res = 1 1 \\
1 1} & \parbox[t]{35mm}{5 + 6\\
...\\
digx = x[0]\\
digx = 5\\
digy = y[0]\\
digy = 6\\
x = x[1:]\\
x = 0\\
y = y[1:]\\
y = 0\\
ds = digx + digy\\
ds = 1 1\\
...\\
res = 1 1 \\
1 1} \\ \\
 \hline
\end{tabular}
\end{center}
\caption{Examples of various CoT demonstrating design choices. Our best and most context efficient is in the \textbf{top left} which excludes code and natural language. It also maintains a \textit{locality} in the context, with digits from $x$ and $y$ being stripped away instead of requiring the model to attend across an increasingly large context to the original input arguments.}
\end{table*}

\end{document}